%% file: Limmer_ICMLA2016.tex
\documentclass[10pt,conference,a4paper]{IEEEtran}

\ifCLASSOPTIONcompsoc
  \usepackage[nocompress]{cite}
\else
  \usepackage{cite}
\fi

\ifCLASSINFOpdf
\else
\fi
\usepackage{graphicx}
\usepackage{amsmath,amssymb}
\usepackage{color}
\usepackage{pgfplots}
\usepackage{subfigure}
\usepackage[utf8]{inputenc}
\usepackage[T1]{fontenc}
\usepackage{lmodern}
\usepackage{wrapfig}
\usepackage{stfloats}
\usepackage{booktabs}
\usepackage{url}
\usepackage{caption}

\usetikzlibrary{external}
\usetikzlibrary{pgfplots.groupplots}

\input{Macros}

\begin{document}

\title{Infrared Colorization Using Deep Convolutional Neural Networks}

\author{\IEEEauthorblockN{Matthias Limmer\IEEEauthorrefmark{1}, Hendrik P.A. Lensch\IEEEauthorrefmark{2}}
\IEEEauthorblockA{\IEEEauthorrefmark{1}Daimler AG, Ulm, Germany}
\IEEEauthorblockA{\IEEEauthorrefmark{2}Department of Computer Graphics, Eberhard Karls Universität, Tübingen, Germany}
}

\maketitle

\input{chapters/abstract}


%
\IEEEpeerreviewmaketitle

\input{chapters/introduction}
\input{chapters/relwork}
\input{chapters/approach}
\input{chapters/experiments}
\input{chapters/discussion}
\input{chapters/summary}

\input{figures/limit}

\input{chapters/acknowledgments}

\bibliographystyle{IEEEtran}
\bibliography{bib}

\end{document}

%% file: Macros.tex
\newcommand{\nallimages}{38.495 }

\newcommand{\nevalimages}{800 }

\newcommand{\ntrainimages}{32.795 }

\newcommand{\itex}{I_{1}'}
\newcommand{\imean}{I_{\mu}}
\newcommand{\tmean}{T_{\mu}}
\newcommand{\emean}{E_{\mu}}
\newcommand{\ivar}{I_{\sigma}}
\newcommand{\idet}{I_{h}}
\newcommand{\relu}{\mathrm{ReLU}}

\newcommand{\argmin}{\mathrm{argmin}}
\newcommand{\roi}{\mathit{roi}}

\newcommand{\topo}[3]{\texttt{topo-#1-#2-#3}}
\newcommand{\topobp}[3]{\texttt{topo-#1-#2-#3-bp}}
\newcommand{\roiS}{46}
\newcommand{\roiM}{68}
\newcommand{\roiL}{98}

%% file: chapters/abstract.tex
\begin{abstract}

This paper proposes a method for transferring the RGB color spectrum to near-infrared (NIR) images using deep multi-scale convolutional neural networks. 
A direct and integrated transfer between NIR and RGB pixels is trained. 
The trained model does not require any user guidance or a reference image database in the recall phase to produce images with a natural appearance. 
To preserve the rich details of the NIR image, its high frequency features are transferred to the estimated RGB image.
The presented approach is trained and evaluated on a real-world dataset containing a large amount of road scene images in summer. 
The dataset was captured by a multi-CCD NIR/RGB camera, which ensures a perfect pixel to pixel registration. 

\end{abstract}

%% file: chapters/introduction.tex
\section{Introduction}
In advanced driver assistance systems, cameras are often used as a sensor for object detection and augmentation (e.g. to alarm the driver of obstacles or possible threats). 
Near-infrared cameras have two advantages over regular RGB cameras. 
First, color or infrared filters are not applied to the sensor, thus not diminishing its sensitivity. 
Second, infrared light beams, which are invisible to the human eye, can be used to illuminate the scene in low light conditions without blinding other road users.
The NIR images, produced by cameras without color filters, are grayscale.
Since the IR-cut filter has been removed, they possess an appearance different from images with an IR-cut filter.
Using such images in augmenting systems decreases the user acceptance, because their look does not agree with human cognition~\cite{Toet2012}.
It is more difficult for users to orientate on images that lack color discrimination or contain wrong colors.
Integrating a second sensor only for display purposes increases the size of the hardware components and the price of the final product.
For this reason, transforming the NIR image into a natural looking RGB image, like in Fig.~\ref{fig:eyecatcher}, is desirable.\\
\input{figures/eyecatcher}

Transforming a grayscale NIR image into a multichannel RGB image is closely related to \emph{Image Colorization}, where regular grayscale images are colorized, and \emph{Color Transfer}, where color distributions are transferred from one RGB image to another.
Both techniques, however, are not simply applicable for colorizing NIR images. 
They often contain multiple cues, including various optimization, feature extraction and segmentation algorithms, and have certain prerequisites.
Colorization, for example, leverages the fact, that the luminance is given by the grayscale input, and therefore only estimates the chrominance.
NIR colorization requires estimating both the luminance and the chrominance.
On the other hand, color transfer methods are often tailored to transform multi-channel input into multi-channel output.
The reduced dimensionality of single-channel NIR images renders many color transfer methods ineffective because they often require inter-color distinction to produce reasonable results.

This paper proposes an integrated approach based on deep-learning techniques to perform a spectral transfer of NIR to RGB images. 
A deep multi-scale \emph{Convolutional Neural Network} (CNN) performs a direct estimation of the low frequency RGB values. 
A postprocessing step that filters the raw output of the CNN and transfers the details of the input image to the final output image is the only additional cue in the proposed approach. 
Sophisticated neural network architectures with a dedicated bypass path are trained on sunny summer rural road scenes.
The images are from a specialized mulit-CCD camera providing pixel-to-pixel registered NIR and RGB images. 
Extensive numerical experiments demonstrate significantly better results, compared to existing colorization and color transfer methods, and the potential of the proposed approach.

%% file: figures/eyecatcher.tex
\begin{figure}[t]
\captionsetup{font=scriptsize}
\centering
\includegraphics[width=0.24\textwidth]{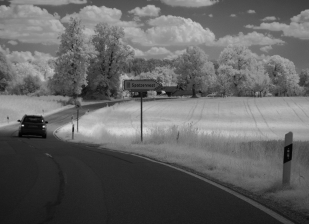}%
\includegraphics[width=0.24\textwidth]{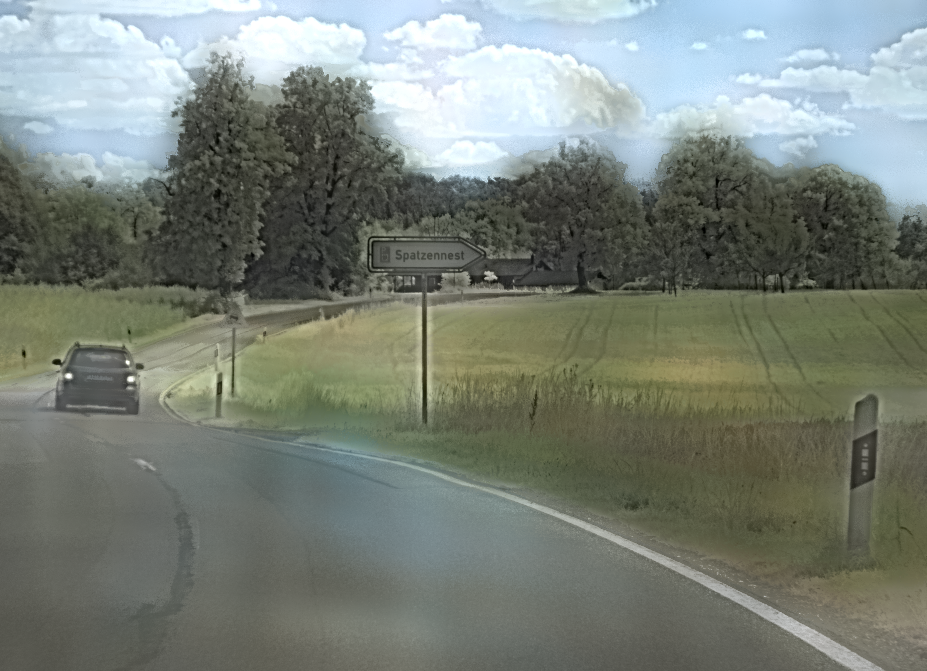}
\caption{An NIR image (Left) is colorized (Right) by the approach described in this paper. Best viewed in color.}
\label{fig:eyecatcher}
\end{figure}

%% file: chapters/relwork.tex
\section{Related Work}
\label{sec:relwork}

Color transfer methods analyze the color distribution of an input image and fit them to a target color distribution, taken from one or more target images.
This can be done globally for the whole image \cite{Reinhard2001,Pitie2007} or locally for image patches or segments 
\cite{Wang2011,Shih2013,Laffont2014}. 
Especially the approaches of~\cite{Shih2013}~and~\cite{Laffont2014} are showing impressive results in mapping day images to night images \cite{Shih2013} or summer images to winter images \cite{Laffont2014}. 
Both approaches determine color transfers for small input patches by finding similar image patches in a sample image pair, which conducts a desired transformation (e.g. frames from a time-lapse video at different times of day). 

Colorization methods are used for transferring colors to grayscale images. 
Typical approaches need to perform three main steps. 
First, the image is segmented into regions that are supposed to receive the same color. 
Second, for each region a target color palette is retrieved. 
Third, the retrieved color palettes for each region are used to determine their respective chrominance. 

Approaches \cite{Levin2004,Yatziv2006,Sheng2011} use \emph{scribbles}\footnote{Scribbles are colored strokes on an image, which mark roughly which area receives which color. They are generally given by a user.} to determine the color palette and the coarse region segmentation. 
Other approaches, such as \cite{Charpiat2008,Pang2013,Deshpande2015}, do not rely on user guidance.
In these approaches, patches of an input image are matched to patches of a reference image or reference colorization by using feature extraction and matching.
This operates fully automatically if fitting target images are given.
However, it is also the greatest drawback of those approaches since a fitting reference image database and retrieval mechanism needs to be implemented.

The approach of~\cite{Cheng2015} uses a \emph{Multi Layer Perceptron} (MLP) to colorize natural images. 
They exploit \emph{Scene Labels} of the used \emph{SUN} dataset \cite{Patterson2012} as an additional feature as well as extracted \emph{DAISY} features \cite{Tola2008} to perform better class specific colorizations.
This approach performs a fully automatic and integrated colorization, but requires scene labels as an input and these are generally not provided for arbitrary sets of images.

Recent approaches~\cite{Dahl2016,Iizuka2016,Larsson2016,Zhang2016} leverage deep CNNs for automatic image colorization.
While~\cite{Dahl2016}~and~\cite{Iizuka2016} train their networks to directly estimate chrominace values, \cite{Larsson2016}~and~\cite{Zhang2016} quantize the chrominance space into discrete colors and perform a logistic regression.
\cite{Dahl2016,Larsson2016}~and~\cite{Zhang2016} initialize their networks with publicly available pre-trained models and adapt them to perform the colorization task. 
Contrary to~\cite{Iizuka2016}~and~\cite{Zhang2016}, who introduce their own network topologies, \cite{Dahl2016}~and~\cite{Larsson2016}~append \emph{Hypercolumns}~\cite{Hariharan2015} to their topologies.
All these approaches show that deep CNNs are suitable to perform automatic image colorizations.
Due to their design of estimating the chrominance only, they are not directly applicable to colorize NIR images.

The approach proposed in this paper uses CNNs to perform an automatic integrated colorization from a single channel NIR image.
Chrominance as well as luminance is reliably inferred for the majority of natural objects in a summerly road scenery setup.
Contrary to previous approaches, though, additional complex processing cues or hand-crafted features, like scene labels, are not utilized.

The remainder of the paper is structured as follows: Section~\ref{sec:approach} describes the proposed approach, while extensive experiments are conducted in Section~\ref{sec:experiments}. 
The results are discussed in Section~\ref{sec:discussion} and concluded in Section~\ref{sec:conclusion}.

%% file: chapters/approach.tex
\section{Approach}\label{sec:approach}
The recent increase in the computational power of general purpose GPUs resulted in larger CNN architectures surpassing state-of-the-art performances in various classification tasks \cite{Simonyan2014,He2015,Szegedy2015,Badrinarayanan2015,Long2015}. 
This indicates that inference problems such as image colorization can reach superior performance if deep CNNs are used.

This paper proposes a method using deep multi-scale CNNs to colorize infrared images. 
The architecture is inspired by \cite{Simonyan2014} and combines it with the multi-scale scheme from \cite{Farabet2013}. 
While the training of the network is patch-based, the inference is image-based using techniques from \cite{Giusti2013}~and~\cite{Thom2016} to preserve the image resolution.

\subsection{Deep Multiscale CNNs}
A recent trend for CNNs is the usage of many convolution layers with small convolution kernels and relatively few pooling layers. 
This increases the total amount of non-linearities in the network.
Furthermore, the computational complexity of each convolution layer is decreased due to small kernel sizes \cite{Simonyan2014}. %
To increase the scale invariance of the approach proposed in this paper multiple scales~\cite{Farabet2013} of the same input data are processed concurrently and lastly combined to one output (e.g. by a \emph{fully connected} layer).

\input{figures/detail}

\subsection{Framework}\label{sec:framew}
The proposed approach consists of three steps: 
First, preprocessing is performed to build a normalized image pyramid. 
Second, the color is inferred by using a CNN. 
Third, postprocessing is completed by filtering the raw output of the network and transferring the details from the input image.
A schematic overview of all processing steps is depicted in Fig.~\ref{fig:detail}.

\subsection{Preprocessing}
The preprocessing component is a two-step approach. 
First, an image pyramid of $n_l$ levels is constructed. 
Every level reduces each image dimension by a factor of $0.5$. 
Second, all pyramid levels are normalized to zero mean unit variance in a local neighborhood. 
Applied to all pixels of an image, this produces an image of enhanced texture.
We denote the normalized image in the first level of the image pyramid by $\itex$. 
Byproducts of the normalization are an image of local standard deviations $\ivar$ and a mean filtered image $\imean$. 
While $\imean$ can be considered the low frequency component of image $I$, so the Hadamard product of $\itex \circ \ivar = \idet$ can be considered the high frequency component of $I$.

\subsection{Inference}
Each level of the input image pyramid is fed to its own branch of the neural network. 
The branches are fused in a final fully-connected layer, which is also the output layer of the network. 
Though each branch is structurally identical, the corresponding layers do not share their weights with each other. 
Each branch overall consists of $n_c$ convolution layers and $n_p$ max-pooling layers. 
The pooling layers are distributed between the convolution layers so that each \emph{convolution layer block} has the same amount of convolution layers. 
The activation function of the convolution layers is the $\relu$ function: $\relu(x) = \max(0,x)$. 
The size of the filter bank $n_f$ is the same inside each convolution layer block and is doubled after each pooling layer. 
The kernel size of the convolution kernels $n_k$ is the same in all convolution layers.  
A peculiarity of the proposed network architecture is an optional bypass connection of the corresponding values of $\imean$ to the final fully connected layer.
Note that a patch-based application requires an extraction of the correctly sized input patches $\roi_i$ for each scale.
This is due to the reduction of resolution by using \emph{valid convolutions}\footnote{Valid convolutions do not pad the image before convolving.} and pooling layers.

\subsection{Postprocessing}
The raw output of the inference step $\emean$ shows visible noise, caused by inaccurate pixel-wise estimations.
The subsampling property of the pooling layers combined with the correlation property of the convolution layers amplify this effect.
Since source pixel locations of the subsampled and convolved coherent output feature maps are not adjacent in the original image (c.f.~\cite{Giusti2013,Thom2016}), 
evaluating every pixel position results in an interleaving of these separate coherent output maps. 
If those maps differ in local regions, the interleaving introduces a checkerboard pattern, as can be seen in Fig.~\ref{fig:bfilt:raw}. 
This \emph{coherence gap} $s_{\pi}$ between pixels of coherent maps can be calculated by building a product of all strides in the neural network $s_{\pi} = \prod{s_{l,i}}$ where ${s_{l,i}\in S_l}$ defines the strides of all layers in scale $l$. 
For a network, which contains three $2\times 2$ pooling layers, the coherence gap is $s_{\pi} = 2^3 = 8$ in both image dimensions. 
Postprocessing is necessary to remove this incoherence and recover the lost details.

The postprocessing consists of two steps: 
First, the raw output after the inference $\emean$ step is filtered by a \emph{Joint Bilateral Filter} using the \emph{Bilateral Grid} \cite{Chen2007}.
Second, the filtered image is augmented by the high frequency details $\idet$ from the input image to produce the final colorized image.

The joint bilateral filter using Gaussians has two major parameters: the spatial domain standard deviation $\sigma_{g}$ and the range domain standard deviation $\sigma_{f}$. 
While $\sigma_{g}$ steers the spatial size of the blur, $\sigma_{f}$ steers the sensitivity to edges in the filtering process. 
A large $\sigma_{g}$ increases the area of the blur, while a small $\sigma_{g}$ does not reduce the noise. 
Likewise, a small $\sigma_{f}$ increases the edge sensitivity and a large $\sigma_{f}$ increases the smoothing effect. 
The joint bilateral filter also utilizes a guidance channel on which the range domain filter is computed.
In the proposed approach, the input image $I$ is used as the guidance channel, because it contains less noise than the raw network output and all necessary surface edges to conduct an edge aware smoothing.
The result after the filtering step is displayed in Fig.~\ref{fig:bfilt:filt}.

Compared to the raw output $\emean$, object contours and edges are clearly visible, but the textures of the surfaces are still missing.
By using the detail component of the input image $\idet$, textures can be partially recovered (see Fig~\ref{fig:bfilt:det}). 

\input{figures/bfilteronly}

%% file: figures/detail.tex
\begin{figure*}[t]
\captionsetup{font=scriptsize}
\centering
\vspace{0.5\baselineskip}%
\includegraphics[width=0.98\textwidth]{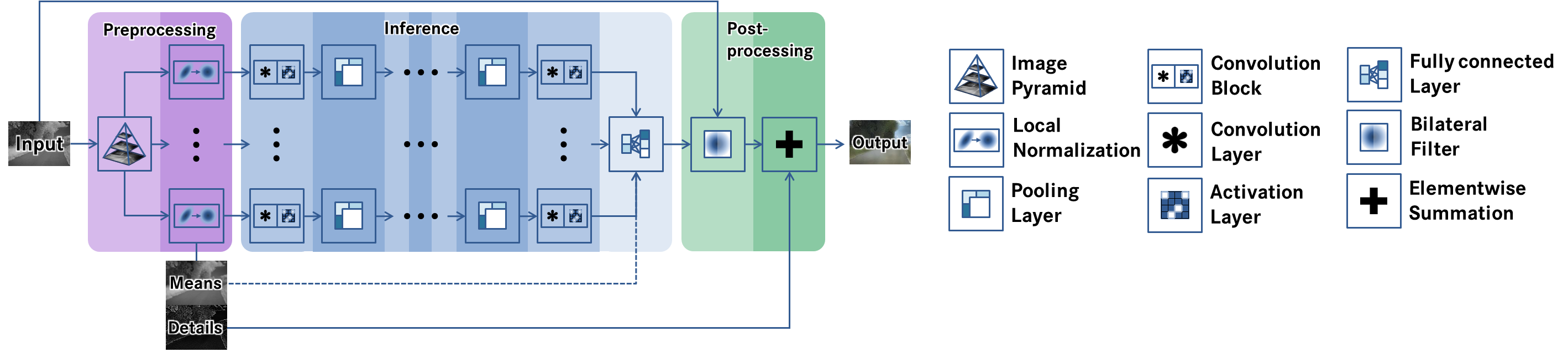}
\caption{A detailed view of all processing steps of the proposed approach: 
The colored blocks denote the various processing steps. 
The arrows indicate a data flow between the processing units. 
All processing units are depicted in a legend on the right. Convolution and activation layers are not explicitly used, because they are included in the convolution blocks. 
Each block consists of a fixed amount of convolution and activation layers. 
The amount of pyramid levels can variable and therefore also the amount of branches in the inference component. 
}\label{fig:detail}
\end{figure*}

%% file: figures/bfilteronly.tex
\begin{figure}
\captionsetup{font=scriptsize}
\centering
\includegraphics[width=0.16\textwidth]{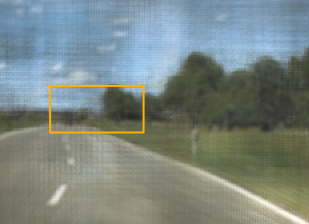}%
\includegraphics[width=0.16\textwidth]{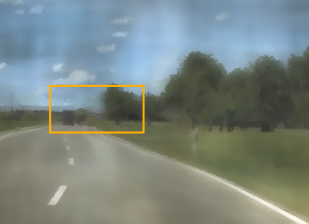}%
\includegraphics[width=0.16\textwidth]{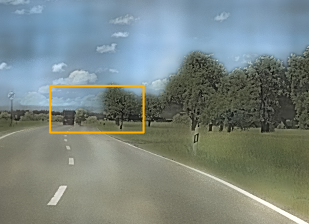}\\
\vspace{-0.5\baselineskip}%
\subfigure[\label{fig:bfilt:raw}Raw Output]{%
\includegraphics[width=0.16\textwidth]{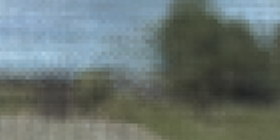}}%
\subfigure[\label{fig:bfilt:filt}Bilateral Filtered]{%
\includegraphics[width=0.16\textwidth]{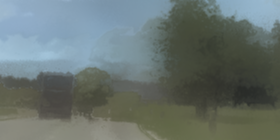}}%
\subfigure[\label{fig:bfilt:det}With Details]{%
\includegraphics[width=0.16\textwidth]{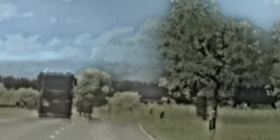}}
\caption{
Results of the postprocessing steps. 
The checkerboard pattern is clearly visible in the raw network output (a). 
It is filtered by a bilateral filter (b) to remove the noise. 
After adding the detail image, textures and distinct object boundaries are visible (c).
}\label{fig:bfilteronly}
\end{figure}

%% file: chapters/experiments.tex
\section{Experiments}\label{sec:experiments}
In the following, the approach proposed in this paper is trained and evaluated on real-world images. 
Data acquisition has been performed using a two-CCD camera\footnote{jAi AD-080CL: \url{http://www.jai.com/en/products/ad-080cl}}. 
The sensor splits NIR and RGB wavelengths and distributes them to dedicated CCDs.
This ensures pixel to pixel registration and temporal synchronization between the channels.
Due to the application scope of the approach proposed in this paper a dataset was assembled accordingly.
Video sequences were recorded during several sunny summer days in a time range of one month resulting in approximately 5\,h of video material with 30 frames per second.
These sequences have then been equally sampled into \nallimages image pairs and show a variation of road scenes, incorporating rural areas as well as highways.
The target RGB images are white balanced and demosaiced. 
Finally, \ntrainimages image pairs were used for training and \nevalimages for evaluation.
The set of image pairs for evaluation is disjoint to those for training taken from different recording days and showing different tracks.
The native resolution of the sensor is $1024\times768$ pixels with a bit-depth of 10 bits per pixel.
Fig.~\ref{fig:dataset} shows various example image pairs from the dataset.
\input{figures/dataset}

\subsection{Color Transfer and Colorization}\label{sec:traditional}
Fig.~\ref{fig:traditional} shows the results of various color transfer and colorization methods applied on one example from our dataset Fig.~\ref{fig:trad:inp}.
For algorithms Fig.~\ref{fig:trad:rei} - Fig.~\ref{fig:trad:shi} the corresponding RGB channel Fig.~\ref{fig:trad:tar} was used as the target image.
Note that Fig.~\ref{fig:trad:tar} would not be present in realistic applications.

Fig.~\ref{fig:trad:lev} displays a user-guided colorization from~\cite{Levin2004} and Fig.~\ref{fig:trad:des} an automatic colorization from~\cite{Deshpande2015}. 
Both approaches are able to colorize the sky, but fail for trees and grass. 
This is mainly caused by not estimating the correct luminance.

Fig.~\ref{fig:trad:rei},~\ref{fig:trad:pit} and~\ref{fig:trad:shi} show the results of two global~\cite{Reinhard2001,Pitie2007} and one local color transfer method~\cite{Shih2013}. 
These approaches are not able to transfer different colors to different objects at all. 
They are apparently not suitable for colorizing NIR images.
For comparison, the result of the approach proposed in this paper is shown in Fig.~\ref{fig:trad:our}. 
Note that this colorization was performed without the knowledge of the target image.
\input{figures/traditional}

\input{figures/topos}

\subsection{Network Topologies}\label{sec:topoex}
Not all parameters mentioned in Section~\ref{sec:framew} are evaluated.
The kernel size $n_k$ of the convolution layers was fixed to a small size of $3\times 3$ pixels and the kernel size of the max pooling to $2\times 2$ pixels.
This is chosen analogous to~\cite{Simonyan2014}, who discovered that a stack of convolution layers with small kernels benefits the classification accuracy greatly compared to singular convolution layers with big kernels.
To counteract the increasing computational complexity and memory consumption of the multi-scale analysis, the filter bank of the first convolution block $n_{f_1}$ is set to a fixed amount of 16 filters and doubled after each pooling layer.
That leaves the amount of convolution layers $n_c$, pooling layers $n_p$ and scales $n_l$ variable. 
We further examine the effect of bypassing the mean image $\imean$ as an input to the final fully connected layer. 
The evaluated configurations are summarized in Table~\ref{tab:topos}. 
Parameters $n_c$ and $n_p$ are chosen according to the following criteria. 
Better results are not expected from configurations smaller than \topo{1}{9}{2}. 
Configurations \topo{*}{12}{3} are deeper versions of \topo{*}{9}{2} with one additional pooling layer and convolution block per scale. 
Additionally, intermediary configurations \topo{*}{8}{3} containing 3 pooling layers and 4 convolution blocks with 2 convolution layers each were investigated. 
The input patch size of \topo{*}{8}{3}, $\roi_i=\roiM$ lies between those of \topo{*}{9}{2}, $roi_i=\roiS$ and \topo{*}{12}{3}, $\roi_i=\roiL$.

All network topologies were trained with a similar scheme.
The trainable parameters $\Theta$ are initialized from a Gaussian random distribution. 
Stochastic gradient descent using the backpropagation algorithm~\cite{LeCun1998b} is performed to minimize the \emph{mean squared error} (MSE) between the pixel estimates $\mathcal{F}(p_{i}',\Theta)$ of the $i^{\mathrm{th}}$ normalized pixel $p_{i}'$ and the corresponding pixels $q_i$ of the mean filtered RGB image $\tmean$:
\begin{equation}
\underset{\Theta}{\argmin}\frac{1}{D}\sum_{i\in D}{||\mathcal{F}(p_{i}',\Theta) - q_i ||^2}
\end{equation}
where $D$ describes the number of pixels in the dataset. 
In each epoch, patches of multiple random pixel locations from a random subset of images are extracted and fed to the network.
The initial learning rates $\eta$ for each experiment were chosen empirically by performing mini-trainings and choosing the best performing $\eta$.
Then full trainings of 10000 epochs were conducted with the selected learn rates using a linear annealing and a momentum of $0.9$. 

\input{figures/iqms}
The topologies from Table~\ref{tab:topos} have been evaluated with respect to the RMSE and \emph{S-CIELAB}~\cite{Zhang1996} image quality measures. 
The S-CIELAB is a subjective measure, designed for measuring the image quality from a human perspective.
The raw network outputs of the 800 images from the evaluation dataset are individually evaluated by both measures before any postprocessing has been applied.
Table~\ref{tab:iqms} displays the average values and standard deviations of the evaluations for each topology.
Fig.~\ref{fig:depth} shows the evaluation results of the RMSE graphically.
The solid lines display the medians and the transparent area their respective interquartile distance.  
The best performing network architecture for both measures is \topobp{3}{12}{3}.
It has the biggest input patch size, 3 scales and the bypass path. 
Although \topobp{1}{12}{3} is behaving otherwise, network architectures tend to perform better if their input patch size increases.
\input{figures/depth}

This also seems to be the case for topologies with more scales.
Additional topologies (\topo{[1..5]}{9}{2} and \topobp{[1..5]}{9}{2}) were trained to examine the influence of the amount of scales to the performance. 
Fig.~\ref{fig:scales} shows that an increase of scales is beneficial to the performance up to a certain threshold. 
More than 4 scales result in almost no improvement (cf. \topo{5}{9}{2}) or even slight loss of performance (cf. \topobp{5}{9}{2}). 
The reason for that behavior is the final resolution of scale 5.
From an initial resolution of $1024 \times 768$ pixels of the input image, scale 5 has a resolution of $64 \times 48$ pixels. 
With a patch size $\roi=\roiS$, the image of scale 5 is almost completely contained by the patch. 
In that case, scale 5 cannot provide any useful information, which benefits a distinguished inference for different pixel locations.
\input{figures/scales}

A positive effect, though, can be recognized for all topologies when using the bypass path. 
It seems that the mean filtered image $\imean$ serves as a prior to $\tmean$, because it contains information that is not present in the normalized input images.

\input{figures/bfiltermatrix}
\input{figures/bfilter_scielab_graph}
\subsection{Postprocessing}\label{sec:postex}
In the postprocessing step, the parameterization of the bilateral filter has a great impact on the final visual appearance. 
Fig.~\ref{fig:bfiltermatrix} shows the visual influence of various instances of $\sigma_{g}$ and $\sigma_{f}$ after details have been added to the estimated image. 
The column on the left (all with $\sigma_g = 5$) still shows strong noise deriving from the coherence gap.
The column on the right (all with $\sigma_g = 65$), however, shows a severe \emph{color bleeding} effect resulting from the great spatial blur of the bilateral filter.
The range parameter $\sigma_f$, though, appears to have a minor effect on the final result. 
This is reflected in Fig.~\ref{fig:bfiltergraph}, which displays the average S-CIELAB error over the evaluation dataset for \topobp{3}{12}{3} in relation to the parameters of the bilateral filter.
The average error of $\emean$ (dashed line) is increased by adding the detail layer $\idet$ (dotted line), since and addition of details was not considered in the training of the CNN.
The bilateral filter, though, is able to reduce this increased error, by choosing $\sigma_g$ and $\sigma_f$ wisely.
Fig.~\ref{fig:bfiltergraph} shows that small $\sigma_f$ perform better, but keep the error in the same range for reasonable $\sigma_g$.
Values of $\sigma_g \leq 17$ are mainly smoothing the noise induced by the coherence gap for this topology.
Values of $\sigma_g > 17$ start to have increasingly negative effects, because the smoothing of bigger areas result in a color bleeding effect.
In these cases the error surpasses the error of $\emean+\idet$.
\input{figures/eyecatcher2}

%% file: figures/dataset.tex
\begin{figure*}
\captionsetup{font=scriptsize}
\centering
\includegraphics[width=0.24\textwidth]{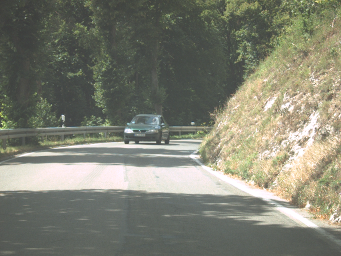}%
\includegraphics[width=0.24\textwidth]{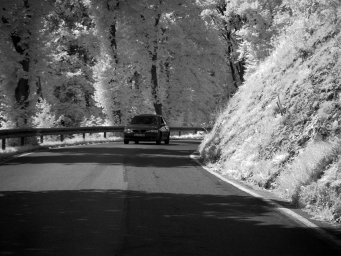}\ \ 
\includegraphics[width=0.24\textwidth]{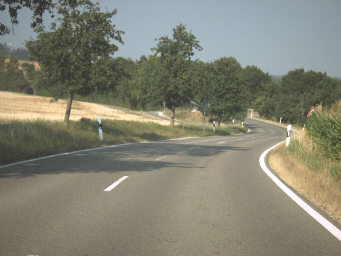}%
\includegraphics[width=0.24\textwidth]{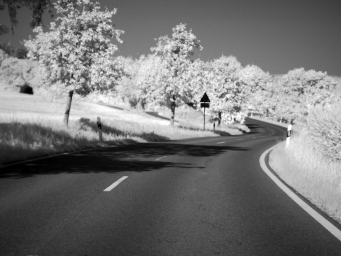}\\
\includegraphics[width=0.24\textwidth]{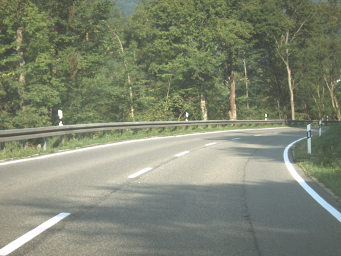}%
\includegraphics[width=0.24\textwidth]{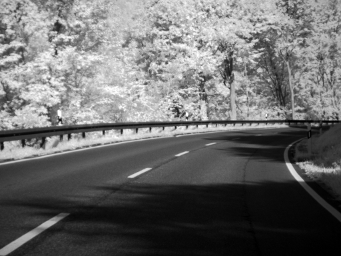}\ \ 
\includegraphics[width=0.24\textwidth]{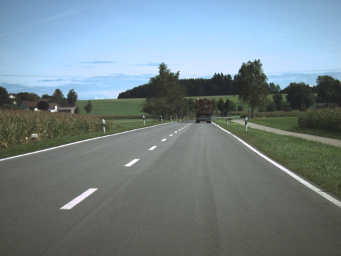}%
\includegraphics[width=0.24\textwidth]{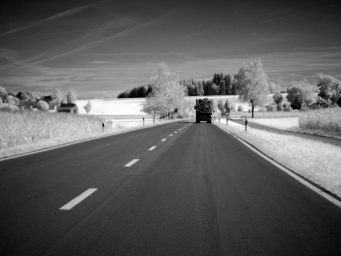}
\caption{
Example image pairs of the used dataset.
The left sides display the RGB and the right sides the corresponding NIR images.
}\label{fig:dataset}
\end{figure*}

%% file: figures/traditional.tex
\begin{figure*}[t]
\captionsetup{font=scriptsize}
\centering
\subfigure[\label{fig:trad:inp}Input]{%
\includegraphics[width=0.245\textwidth]{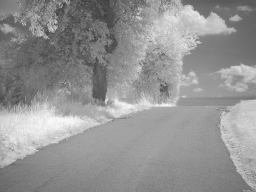}}%
\subfigure[\label{fig:trad:lev}\cite{Levin2004}]{%
\includegraphics[width=0.245\textwidth]{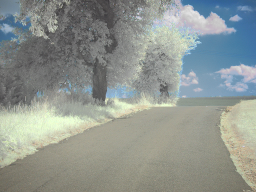}}%
\subfigure[\label{fig:trad:des}\cite{Deshpande2015}]{%
\includegraphics[width=0.245\textwidth]{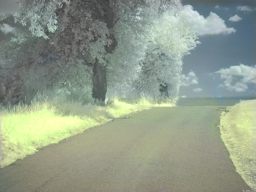}}%
\subfigure[\label{fig:trad:rei}\cite{Reinhard2001}]{%
\includegraphics[width=0.245\textwidth]{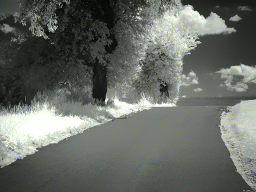}}\\%
\vspace{-0.5\baselineskip}%
\subfigure[\label{fig:trad:pit}\cite{Pitie2007}]{%
\includegraphics[width=0.245\textwidth]{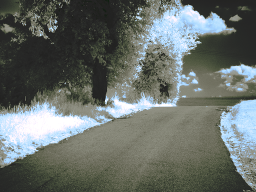}}%
\subfigure[\label{fig:trad:shi}\cite{Shih2013}]{%
\includegraphics[width=0.245\textwidth]{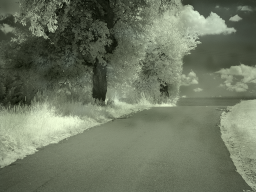}}%
\subfigure[\label{fig:trad:our}Ours]{%
\includegraphics[width=0.245\textwidth]{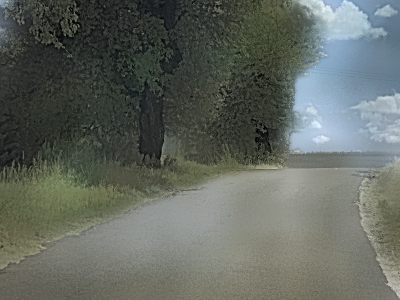}}%
\subfigure[\label{fig:trad:tar}Target]{%
\includegraphics[width=0.245\textwidth]{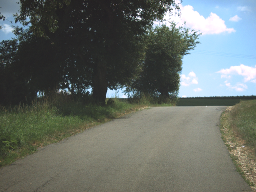}}
\caption{
Colorization and Color Transfer methods applied to an NIR image (a) using the corresponding RGB image (h). 
(b) is the result of \cite{Levin2004}, a user guided colorization method, using scribbles. 
(c) is the result of \cite{Deshpande2015}, an automatic colorization method. 
(d) shows the result of \cite{Reinhard2001}, a simple global color transfer method, while (e) shows a more sophisticated global color transfer from \cite{Pitie2007}. 
(f) is the result of \cite{Shih2013}, a local color transfer method. 
Our result is displayed in (g). 
It was colorized without using (h) as training data.
}\label{fig:traditional}
\end{figure*}

%% file: figures/topos.tex
\begin{table}
\captionsetup{font=scriptsize}
\centering
\begin{tabular}{lccccc}
\toprule
  Name & $n_l$ & $n_c$ & $n_p$ & bypass & $\roi_i$ \\
  \midrule
  \topo{1}{9}{2} & 1 & 9 & 2 & - & \roiS \\
  \topobp{1}{9}{2} & 1 & 9 & 2 & \checkmark & \roiS \\
  \topo{1}{8}{3} & 1 & 8 & 3 & - & \roiM \\
  \topobp{1}{8}{3} & 1 & 8 & 3 & \checkmark & \roiM \\
  \topo{1}{12}{3} & 1 & 12 & 3 & - & \roiL \\
  \topobp{1}{12}{3} & 1 & 12 & 3 & \checkmark & \roiL \\
  \topo{3}{9}{2} & 3 & 9 & 2 & - & \roiS \\
  \topobp{3}{9}{2} & 3 & 9 & 2 & \checkmark & \roiS \\
  \topo{3}{8}{3} & 3 & 8 & 3 & - & \roiM \\
  \topobp{3}{8}{3} & 3 & 8 & 3 & \checkmark & \roiM \\
  \topo{3}{12}{3} & 3 & 12 & 3 & - & \roiL \\
  \topobp{3}{12}{3} & 3 & 12 & 3 & \checkmark & \roiL \\
  \bottomrule
\end{tabular}\\
\vspace{0.5\baselineskip}%
\caption{Network topologies, evaluated in this paper. The \emph{bypass} column indicates topologies, where the values of $\imean$ are bypassed to the fully connected layer.}\label{tab:topos}
\end{table}

%% file: figures/iqms.tex
\begin{table}
\captionsetup{font=scriptsize}
\centering
\vspace{1\baselineskip}%
\begin{tabular}{lr@{\hskip 0.5\tabcolsep}c@{\hskip 0.5\tabcolsep}lr@{\hskip 0.5\tabcolsep}c@{\hskip 0.5\tabcolsep}l }
\toprule
    \multicolumn{1}{c}{Name} & \multicolumn{3}{c}{RMSE} & \multicolumn{3}{c}{S-CIELAB} \\
    \midrule
    \topo{1}{9}{2} 		& 0.199 			&$\pm$& 0.049 & 13.36 			&$\pm$& 3.71 \\
    \topo{1}{8}{3} 		& 0.196 			&$\pm$& 0.041 & 13.33 			&$\pm$& 3.19 \\
    \topo{1}{12}{3} 	& 0.194 			&$\pm$& 0.043 & 12.58 			&$\pm$& 3.17 \\
    \topobp{1}{9}{2} 	& 0.151 			&$\pm$& 0.030 & 10.29 			&$\pm$& 2.87 \\
    \topobp{1}{8}{3} 	& 0.146 			&$\pm$& 0.027 & 9.65 			&$\pm$& 2.40 \\
    \topobp{1}{12}{3} 	& 0.153 			&$\pm$& 0.029 & 10.36 			&$\pm$& 2.45 \\
    \topo{3}{9}{2} 		& 0.167 			&$\pm$& 0.047 & 11.21 			&$\pm$& 3.63 \\
    \topo{3}{8}{3} 		& 0.166 			&$\pm$& 0.047 & 11.64 			&$\pm$& 3.44 \\
    \topo{3}{12}{3} 	& 0.155 			&$\pm$& 0.047 & 10.76 			&$\pm$& 3.31 \\
    \topobp{3}{9}{2} 	& 0.149 			&$\pm$& 0.036 & 10.48 			&$\pm$& 3.08 \\
    \topobp{3}{8}{3} 	& 0.137 			&$\pm$& 0.040 & 9.57 			&$\pm$& 3.25 \\
    \topobp{3}{12}{3} 	& \textbf{0.130} 	&$\pm$& 0.043 & \textbf{8.88} 	&$\pm$& 3.15 \\
 \bottomrule
 \\    
\end{tabular}\\
\vspace{0.5\baselineskip}%
\caption{The average RMSE and S-CIELAB~\cite{Zhang1996} and their standard deviation for various network topologies. Best performing topology is \topobp{3}{12}{3}.}\label{tab:iqms}
\end{table}

%% file: figures/depth.tex
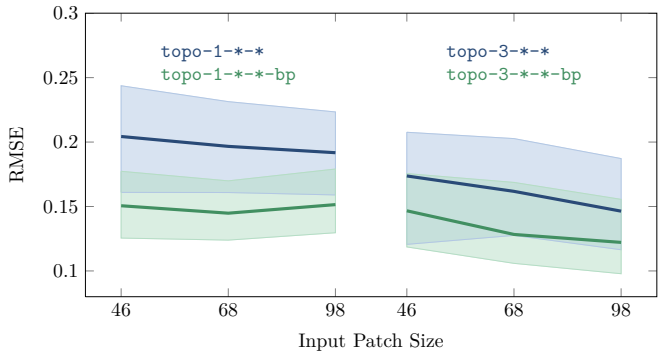
\begin{figure}
\captionsetup{font=scriptsize}
	\tikzsetnextfilename{depth/depth}
	\input{figures/depth.tikz}
	\caption{Medians (solid line) and interquartile range (transparent area) of the RMSE over the evaluation dataset for different network topologies.}
	\label{fig:depth}
\end{figure}

%% file: figures/depth.tikz
%
\definecolor{purple_l}{rgb}{0.83921568627450980392156862745098, 0.72549019607843137254901960784314, 0.92156862745098039215686274509804}%
\definecolor{purple_d}{rgb}{0.61960784313725490196078431372549, 0.36470588235294117647058823529412, 0.81176470588235294117647058823529}%
\definecolor{blue_d}{rgb}{0.16078431372549019607843137254902, 0.29019607843137254901960784313725, 0.46666666666666666666666666666667}%
\definecolor{blue_l}{rgb}{0.69803921568627450980392156862745, 0.78039215686274509803921568627451, 0.89411764705882352941176470588235}%
\definecolor{green_d}{rgb}{0.25490196078431372549019607843137, 0.56078431372549019607843137254902, 0.38039215686274509803921568627451}%
\definecolor{green_l}{rgb}{0.70980392156862745098039215686275, 0.86666666666666666666666666666667, 0.77254901960784313725490196078431}%
\begin{tikzpicture}[scale=0.74]

\begin{axis}[%
width=4in,
height=2in,
unbounded coords=jump,
clip=false,
scale only axis,
xmin=0.5,
xmax=8.5,
xtick={1,2.5,4,5,6.5,8},
xticklabels={\roiS,\roiM,\roiL,\roiS,\roiM,\roiL},
xlabel style={yshift=-1cm},
ymin=0.08,
ymax=0.3,
ylabel={RMSE},
xlabel style={yshift=1cm},
xlabel={Input Patch Size}
]
\addplot [color=blue_l,solid,forget plot,fill=blue_l,opacity=.5]
 table[row sep=crcr]{1	0.160949505155685\\
 2.5	0.160850936858\\
 4	0.158976202500422\\
 4	0.223367811448898\\
 2.5	0.231353543172439\\
 1	0.243679659306159\\
 1	0.160949505155685\\
};
\addplot [color=green_l,solid,forget plot,fill=green_l,opacity=.5]
 table[row sep=crcr]{1	0.177301748870588\\
2.5	0.169874881481592\\
4	0.17906837442799\\
4	0.129652624231455\\
2.5	0.123925156074577\\
1	0.125522938810342\\
1	0.177301748870588\\
};
\addplot [color=blue_l,solid,forget plot,opacity=1]
 table[row sep=crcr]{1	0.160949505155685\\
 2.5	0.160850936858\\
 4	0.158976202500422\\
 4	0.223367811448898\\
 2.5	0.231353543172439\\
 1	0.243679659306159\\
 1	0.160949505155685\\
};
\addplot [color=green_l,solid,forget plot,opacity=1]
 table[row sep=crcr]{1	0.177301748870588\\
2.5	0.169874881481592\\
4	0.17906837442799\\
4	0.129652624231455\\
2.5	0.123925156074577\\
1	0.125522938810342\\
1	0.177301748870588\\
};
\addplot [color=blue_l,solid,forget plot,fill=blue_l,opacity=.5]
table[row sep=crcr]{5	0.20759455002713\\
6.5	0.202685927547529\\
8	0.18717991229877\\
8	0.116454184837134\\
6.5	0.127579934304342\\
5	0.12063823412335\\
5	0.20759455002713\\
};
\addplot [color=green_l,solid,forget plot,fill=green_l,opacity=.5]
table[row sep=crcr]{5	0.175520637554638\\
6.5	0.168673392312154\\
8	0.155530380869717\\
8	0.0978522650595465\\
6.5	0.105892546348866\\
5	0.118645922964776\\
5	0.175520637554638\\
};
\addplot [color=blue_l,solid,forget plot,opacity=1]
table[row sep=crcr]{5	0.20759455002713\\
6.5	0.202685927547529\\
8	0.18717991229877\\
8	0.116454184837134\\
6.5	0.127579934304342\\
5	0.12063823412335\\
5	0.20759455002713\\
};
\addplot [color=green_l,solid,forget plot,opacity=1]
table[row sep=crcr]{5	0.175520637554638\\
6.5	0.168673392312154\\
8	0.155530380869717\\
8	0.0978522650595465\\
6.5	0.105892546348866\\
5	0.118645922964776\\
5	0.175520637554638\\
};
\addplot [color=blue_d,solid,style=ultra thick]
  table[row sep=crcr]{
  1	0.204318783154711\\
  2.5	0.196653393038035\\
  4	0.19174128764268\\
};
\addplot [color=green_d,solid,style=ultra thick]
  table[row sep=crcr]{
  1	0.150583731238298\\
  2.5	0.144806698230571\\
  4	0.151466159643766\\
};
\addplot [color=blue_d,solid,style=ultra thick]
  table[row sep=crcr]{
  5	0.173700996522813\\
  6.5	0.16174998719679\\
  8	0.146381148313839\\
};
\addplot [color=green_d,solid,style=ultra thick]
  table[row sep=crcr]{
  5	0.146661304291658\\
  6.5	0.128321660248855\\
  8	0.122146595935493\\
};
\node [align=left] at (200,180) {\textcolor{blue_d}{\topo{1}{*}{*}}\\ \textcolor{green_d}{\topobp{1}{*}{*}}};
\node [align=left] at (600,180) {\textcolor{blue_d}{\topo{3}{*}{*}}\\ \textcolor{green_d}{\topobp{3}{*}{*}}};
\end{axis}
\end{tikzpicture}%

%% file: figures/scales.tex
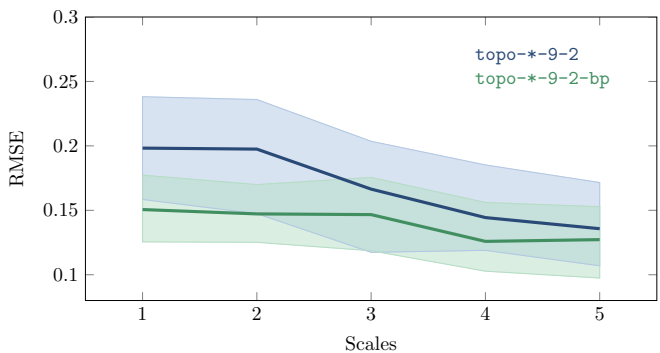
\begin{figure}
\captionsetup{font=scriptsize}
	\tikzsetnextfilename{scales/scales}
	\input{figures/scales.tikz}
	\caption{Medians (solid line) and interquartile range (transparent area) of the RMSE over the evaluation dataset for network topologies \topo{*}{9}{2} and \topobp{*}{9}{2} for different scales.}
	\label{fig:scales}
\end{figure}

%% file: figures/scales.tikz
%
%
\definecolor{purple_l}{rgb}{0.83921568627450980392156862745098, 0.72549019607843137254901960784314, 0.92156862745098039215686274509804}%
\definecolor{purple_d}{rgb}{0.61960784313725490196078431372549, 0.36470588235294117647058823529412, 0.81176470588235294117647058823529}%
\definecolor{blue_d}{rgb}{0.16078431372549019607843137254902, 0.29019607843137254901960784313725, 0.46666666666666666666666666666667}%
\definecolor{blue_l}{rgb}{0.69803921568627450980392156862745, 0.78039215686274509803921568627451, 0.89411764705882352941176470588235}%
\definecolor{green_d}{rgb}{0.25490196078431372549019607843137, 0.56078431372549019607843137254902, 0.38039215686274509803921568627451}%
\definecolor{green_l}{rgb}{0.70980392156862745098039215686275, 0.86666666666666666666666666666667, 0.77254901960784313725490196078431}%
\begin{tikzpicture}[scale=0.74]

\begin{axis}[%
width=4in,
height=2in,
unbounded coords=jump,
clip=false,
scale only axis,
xmin=0.5,
xmax=5.5,
xtick={1,2,3,4,5},
xticklabels={1,2,3,4,5},
xlabel style={yshift=-1cm},
ymin=0.08,
ymax=0.3,
ylabel={RMSE},
xlabel style={yshift=1cm},
xlabel={Scales}
]
\addplot [color=blue_l,solid,forget plot,fill=blue_l,opacity=.5]
 table[row sep=crcr]{1	0.238129051254132\\
 2	0.235949378314597\\
 3	0.203554606960166\\
 4	0.185197682312208\\
 5	0.171548137020143\\
 5	0.106964054733457\\
 4	0.118942619944697\\
 3	0.117385347564286\\
 2	0.147427523534581\\
 1	0.158406732955817\\
 1	0.238129051254132\\
};
\addplot [color=green_l,solid,forget plot,fill=green_l,opacity=.5]
 table[row sep=crcr]{1	0.177301748870588\\
 2	0.169964119032486\\
 3	0.175520637554638\\
 4	0.156102217853554\\
 5	0.152765728491523\\
 5	0.0974853498340908\\
 4	0.102773125573647\\
 3	0.118645922964776\\
 2	0.125192039963657\\
 1	0.125522938810342\\
 1	0.177301748870588\\
};
\addplot [color=blue_l,solid,forget plot,opacity=1]
 table[row sep=crcr]{1	0.238129051254132\\
 2	0.235949378314597\\
 3	0.203554606960166\\
 4	0.185197682312208\\
 5	0.171548137020143\\
 5	0.106964054733457\\
 4	0.118942619944697\\
 3	0.117385347564286\\
 2	0.147427523534581\\
 1	0.158406732955817\\
 1	0.238129051254132\\
};
\addplot [color=green_l,solid,forget plot,opacity=1]
 table[row sep=crcr]{1	0.177301748870588\\
 2	0.169964119032486\\
 3	0.175520637554638\\
 4	0.156102217853554\\
 5	0.152765728491523\\
 5	0.0974853498340908\\
 4	0.102773125573647\\
 3	0.118645922964776\\
 2	0.125192039963657\\
 1	0.125522938810342\\
 1	0.177301748870588\\
};
\addplot [color=blue_d,solid,style=ultra thick]
  table[row sep=crcr]{1	0.198257459452487\\
 2	0.197500069849755\\
 3	0.166437022362538\\
 4	0.144365524760773\\
 5	0.135725754660762\\
};
\addplot [color=green_d,solid,style=ultra thick]
  table[row sep=crcr]{1	0.150583731238298\\
 2	0.147213595755655\\
 3	0.146661304291658\\
 4	0.125886803101705\\
 5	0.127260167981936\\
};
\node [align=left] at (400,180) {\textcolor{blue_d}{\topo{*}{9}{2}}\\ \textcolor{green_d}{\topobp{*}{9}{2}}};
\end{axis}
\end{tikzpicture}%

%% file: figures/bfiltermatrix.tex
{
\begin{figure}
\captionsetup{font=scriptsize}
\centering
\begin{tabular}{c@{}c@{}cl}
\begin{minipage}{0.12\textwidth}
\includegraphics[width=\textwidth]{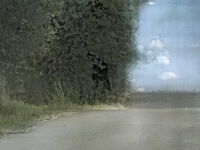} 
\end{minipage} &
\begin{minipage}{0.12\textwidth}
\includegraphics[width=\textwidth]{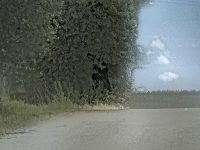} 
\end{minipage} &
\begin{minipage}{0.12\textwidth}
\includegraphics[width=\textwidth]{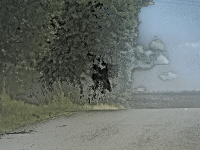} 
\end{minipage} &
{\scriptsize$\sigma_{f}=0.0003$}
\\%
\begin{minipage}{0.12\textwidth}
\includegraphics[width=\textwidth]{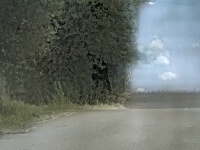} 
\end{minipage} &
\begin{minipage}{0.12\textwidth}
\includegraphics[width=\textwidth]{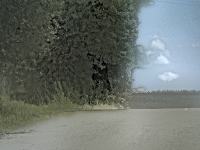} 
\end{minipage} &
\begin{minipage}{0.12\textwidth}
\includegraphics[width=\textwidth]{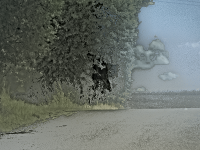} 
\end{minipage} &
{\scriptsize$\sigma_{f}=0.005$}
\\%
\begin{minipage}{0.12\textwidth}
\includegraphics[width=\textwidth]{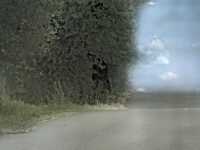} 
\end{minipage} &
\begin{minipage}{0.12\textwidth}
\includegraphics[width=\textwidth]{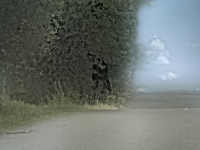} 
\end{minipage} &
\begin{minipage}{0.12\textwidth}
\includegraphics[width=\textwidth]{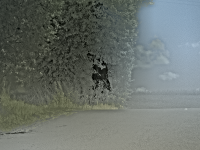} 
\end{minipage} &
{\scriptsize$\sigma_{f}=0.08$}
\\%
{\scriptsize$\sigma_{g}=5$}\ & 
{\scriptsize$\sigma_{g}=17$}\ & 
{\scriptsize$\sigma_{g}=65$} & \\
\end{tabular}\\
\caption{
The effect of different parameterizations of the bilateral filter on the output image of \topobp{3}{12}{3} after adding the details. 
The rows show the effect of different range parameters $\sigma_f$, while the columns show the effect of different spatial parameters $\sigma_g$. 
}\label{fig:bfiltermatrix}
\end{figure}
}

%% file: figures/bfilter_scielab_graph.tex
\begin{figure}[t]
\captionsetup{font=scriptsize}
	\tikzsetnextfilename{bfiltergraph/bfiltergraph}
	\input{figures/bfilter_scielab_graph.tikz}
	\caption{
	The average S-CIELAB errors over all evaluation images of \topobp{3}{12}{3} for the parameters $(\sigma_g,\sigma_f)$ of the bilateral filter after adding the detail layer $\idet$.
	Each graph displays the result of one $\sigma_f$ in relation to various $\sigma_g$.
	For comparison, $\emean$ (dashed line) displays the error of the raw CNN-output and $\emean+\idet$ (dotted line) the error of the raw output after adding the high frequency details.
	}
	\label{fig:bfiltergraph}
\end{figure}
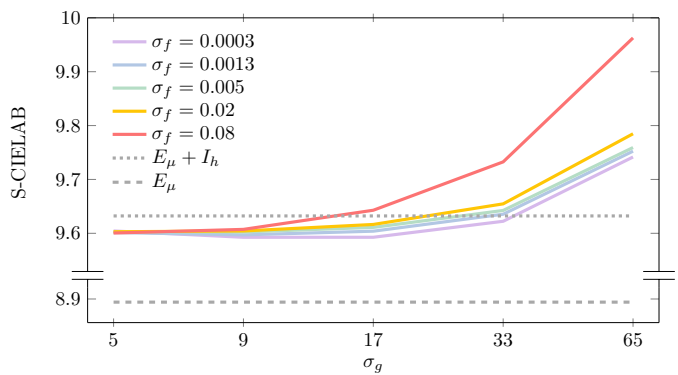

%% file: figures/bfilter_scielab_graph.tikz
%
%
%
\definecolor{mycolor1}{rgb}{0.66,0.66,0.66}%
\definecolor{mycolor2}{rgb}{0.99000,0.4500,0.45}%
\definecolor{mycolor4}{rgb}{0.83921568627450980392156862745098, 0.72549019607843137254901960784314, 0.92156862745098039215686274509804}%
\definecolor{mycolor5}{rgb}{0.69803921568627450980392156862745, 0.78039215686274509803921568627451, 0.89411764705882352941176470588235}%
\definecolor{mycolor6}{rgb}{0.70980392156862745098039215686275, 0.86666666666666666666666666666667, 0.77254901960784313725490196078431}%
\definecolor{mycolor7}{rgb}{1.0,0.77470588235294117647058823529412,0.02060784313725490196078431372549}%
\begin{tikzpicture}[scale=0.74]
\begin{groupplot}[
	group style={
        group name=gplot,
        group size=1 by 2,
        xticklabels at=edge bottom,
        vertical sep=0pt
    },
    width=4in,
	unbounded coords=jump,
	clip=false,
	scale only axis,
	xmin=0.8,
	xmax=5.2,
	xtick={1,2,3,4,5},
	xticklabels={5,9,17,33,65},
	xlabel style={yshift=-1cm},
	ymin=8.8
	ymax=10,
	ylabel={S-CIELAB},
	xlabel style={yshift=1cm},
	xlabel={$\sigma_g$}
]


\nextgroupplot[ymin=9.5,ymax=10,
               ytick={9.6,9.7,9.8,9.9,10},
               axis x line=top, xlabel={},
               x axis line style=-,
               axis y discontinuity=parallel,
               height=1.9in,
               legend pos=north west,
               legend style={draw=none,cells={right} },
			   legend entries={
			   $\sigma_f=0.0003$, 
			   $\sigma_f=0.0013$,
			   $\sigma_{f}=0.005$,
			   $\sigma_f=0.02$,
			   $\sigma_{f}=0.08$,
			   $\emean+\idet$,
			   $\emean$
			   }]
\addlegendimage{no markers, mycolor4, style=ultra thick}
\addlegendimage{no markers, mycolor5, style=ultra thick}
\addlegendimage{no markers, mycolor6, style=ultra thick}
\addlegendimage{no markers, mycolor7, style=ultra thick}
\addlegendimage{no markers, mycolor2, style=ultra thick}
\addlegendimage{no markers, mycolor1, dotted, style=ultra thick}
\addlegendimage{no markers, mycolor1, dashed, style=ultra thick}

\addplot [color=mycolor4,solid,style=ultra thick]
  table[row sep=crcr]{1	9.60436413550928\\
2	9.59260537272803\\
3	9.59264302876814\\
4	9.62231981735671\\
5	9.74163200852397\\
};
\addplot [color=mycolor5,solid,style=ultra thick]
  table[row sep=crcr]{1	9.60141117120487\\
2	9.59710372310539\\
3	9.60392801530586\\
4	9.63542666805341\\
5	9.7524625612545\\
};
\addplot [color=mycolor6,solid,style=ultra thick]
  table[row sep=crcr]{1	9.60275455441209\\
2	9.60242814774505\\
3	9.61100512989608\\
4	9.64243690951536\\
5	9.7592196149799\\
};
\addplot [color=mycolor7,solid,style=ultra thick]
  table[row sep=crcr]{1	9.60291953532197\\
2	9.6043259452247\\
3	9.61644294949952\\
4	9.65467450760204\\
5	9.78478022282766\\
};
\addplot [color=mycolor2,solid,style=ultra thick]
  table[row sep=crcr]{1	9.60051058632261\\
2	9.60730995642805\\
3	9.6428368282001\\
4	9.73260188874787\\
5	9.96274764279958\\
};
\addplot [color=mycolor1,dotted,style=ultra thick]
  table[row sep=crcr]{1	9.6322587429897\\
2	9.6322587429897\\
3	9.6322587429897\\
4	9.6322587429897\\
5	9.6322587429897\\
};

\nextgroupplot[ymin=8.8,ymax=8.95,
               ytick={8.9}, ylabel={},
               yticklabels={8.9},
               axis x line=bottom, 
               x axis line style=-,
               height=0.25in]
\addplot [color=mycolor1,dashed,style=ultra thick]
  table[row sep=crcr]{1	8.88666747374014\\
2	8.88666747374014\\
3	8.88666747374014\\
4	8.88666747374014\\
5	8.88666747374014\\
};

\end{groupplot}
\end{tikzpicture}%

%% file: figures/eyecatcher2.tex
\begin{figure*}
\captionsetup{font=scriptsize}
\centering
\includegraphics[width=0.32\textwidth]{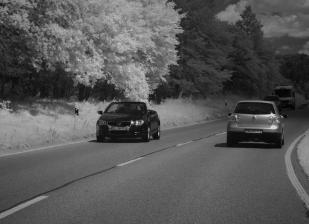}%
\includegraphics[width=0.32\textwidth]{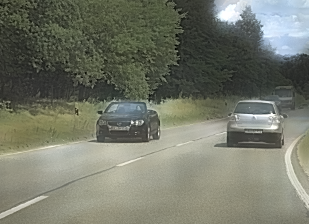}%
\includegraphics[width=0.32\textwidth]{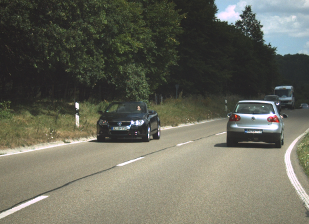}\\
\includegraphics[width=0.32\textwidth]{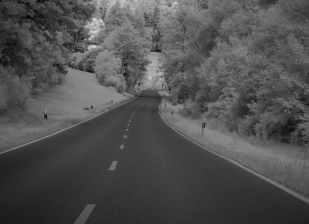}%
\includegraphics[width=0.32\textwidth]{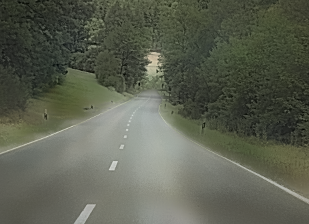}%
\includegraphics[width=0.32\textwidth]{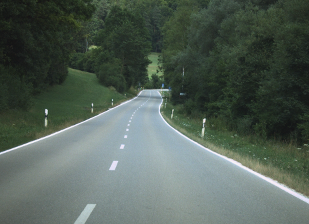}\\
\includegraphics[width=0.32\textwidth]{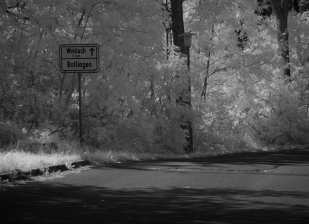}%
\includegraphics[width=0.32\textwidth]{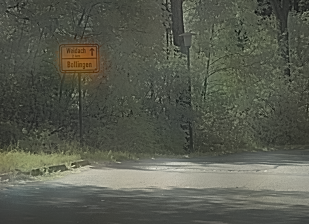}%
\includegraphics[width=0.32\textwidth]{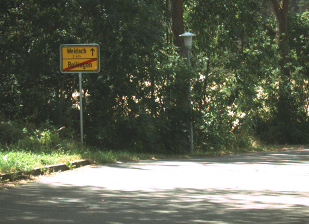}
\caption{
NIR images (Left) are colorized (Middle) in comparison to their target images (Right).
The used topology is \topobp{3}{12}{3} with $\sigma_g = 17$ and $\sigma_f=0.005$.
}\label{fig:eyecatcher2}
\end{figure*}

%% file: chapters/discussion.tex
\section{Discussion}\label{sec:discussion}
The proposed approach is able to colorize NIR images of summerly road scenes fully automatically by leveraging the advantages of a deep neural network.
It cannot reconstruct all information that is not included in single channel NIR images, though. 
Many artificial objects, such as cars, buildings, etc. are falsely colorized because their appearance does not correlate with a specific color (c.f. Fig.~\ref{fig:limit:art}). 
There are also items that are invisible in the NIR image due to the filtering of the RGB wavelengths.
Fig.~\ref{fig:limit:nir} shows an example, where the green light of an LED traffic light is absent from the NIR image.
The induction of model information including additional scene label features might help in some cases, but effects, such as the missing signal of a traffic light, are not recoverable. 

%% file: chapters/summary.tex
\section{Conclusion and Future Work}\label{sec:conclusion}
This paper presented an integrated approach to transfer the color spectrum of an RGB image to an NIR image. 
The transfer is performed by feeding a locally normalized image pyramid to a deep multi-scale CNN, which directly estimates RGB values.
Using the mean filtered input image as an additional input to the final fully connected layer improves the performance greatly.
The resulting raw output is then joint-bilaterally filtered using the input image as a guidance map. 
The details of the input image are added at the end to produce a naturally-colorized output image.
The approach is only failing to colorize objects correctly, where object appearance and color do not correlate.

Future work will incorporate the extension of the training and evaluation dataset to other seasons of the year.
This might result in training distinct colorizers for each of these seasons.
To improve realism of the colorized image textures, future work will also contain the estimation of a correct detail layer.
Changing the loss function to support multi-modal estimations in combination with a semantic segmentation might increase the vibrancy of objects that our algorithm is failing to colorize.

%% file: figures/limit.tex
\begin{figure}
\captionsetup{font=scriptsize}
\centering
\vspace{0.25\baselineskip}%
\subfigure[The color recovery for a truck fails.\label{fig:limit:art}]{%
\includegraphics[width=0.16\textwidth]{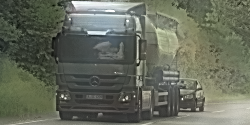}%
\includegraphics[width=0.16\textwidth]{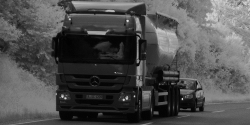}%
\includegraphics[width=0.16\textwidth]{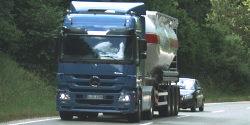}}\\%
\subfigure[The green lamp of an LED traffic light is not perceived by the NIR channel and is therefore not colorized.\label{fig:limit:nir}]{%
\includegraphics[width=0.16\textwidth]{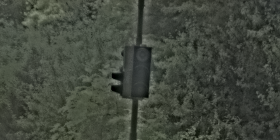}%
\includegraphics[width=0.16\textwidth]{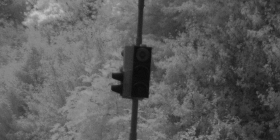}%
\includegraphics[width=0.16\textwidth]{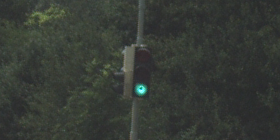}}%
\caption{
Situations, where the approach shows limitations. 
}\label{fig:limit}
\end{figure}

%% file: chapters/acknowledgments.tex
\ifCLASSOPTIONcompsoc
  \section*{Acknowledgments}
\else
  \section*{Acknowledgment}
\fi

Authors would like to thank Markus Thom and Roland Schweiger for valuable advises and input.